\documentclass{ifacconf}

\makeatletter
\let\old@ssect\@ssect % Store how ifacconf defines \@ssect
\makeatother
\usepackage{amsmath,amsfonts,amssymb}
\usepackage{gensymb}
\usepackage{graphicx}      % include this line if your document contains figures
\usepackage{natbib}        % required for bibliography
\usepackage{verbatim}
\usepackage{siunitx}

\usepackage{hyperref}
\usepackage{tabularx}
\usepackage{placeins}
\usepackage{xurl}

\makeatletter
\def\@ssect#1#2#3#4#5#6{%
  \NR@gettitle{#6}% Insert key \nameref title grab
  \old@ssect{#1}{#2}{#3}{#4}{#5}{#6}% Restore ifacconf's \@ssect
}
\makeatother

\usepackage[acronym]{glossaries}
\newacronym[plural=AUVs,firstplural=Autonomous Underwater Vehicles (AUVs)]{auv}{AUV}{Autonomous Underwater Vehicle}
\newacronym[plural=ASVs,firstplural=Autonomous Surface Vehicles (ASVs)]{asv}{ASV}{Autonomous Surface Vehicle}
\newacronym[plural=ROVs, firstplural=Remotely Operated Vehicles (ROVs)]{rov}{ROV}{Remotely Operated Vehicle}
\newacronym[plural=USVs, firstplural=Unmanned Surface Vehicles (USVs)]{usv}{USV}{Unmanned Surface Vehicle}
\newacronym{uuv}{UUV}{Unmanned Underwater Vehicle}
\newacronym[]{mavlink}{MAVLink}{Micro  Aerial Vehicle Link}
\newacronym[]{ros}{ROS}{Robot Operating System}
\newacronym[plural=UVs,firstplural=Unmanned Vehicles (UVs)]{uv}{UV}{Unmanned Vehicle}
\newacronym[plural=UWSNs,firstplural=Underwater Wireless Sensor Networks (UWSNs)]{uwsn}{UWSN}{Underwater Wireless Sensor Network}
\newacronym[plural=PCBs,firstplural=Printed Circuit Boards (PCBs)]{pcb}{PCB}{Printed Circuit Board}
\newacronym[plural=GNSSs,firstplural=Global Navigation Satellite System (GNSSs)]{gnss}{GNSS}{Global Navigation Satellite System}
\newacronym{imu}{IMU}{Inertial Measurement Unit}
\newacronym{lan}{LAN}{Local Area Network}
\newacronym{gnc}{GNC}{Guidance Navigation and Control}
\newacronym{los}{LoS}{Line-of-Sight}
\newacronym{slam}{SLAM}{Simultaneous Localisation and Mapping}
\newacronym{lidar}{LiDAR}{Light Detection and Ranging}
\newacronym{sbl}{SBL}{Short Baseline}
\newacronym{usbl}{USBL}{Ultra-Short Baseline}
\newacronym{dvl}{DVL}{Doppler Velocity Logger}
\newacronym{rtk}{RTK}{Real-Time Kinematics}
\newacronym{ekf}{EKF}{Extended Kalman Filter}
\newacronym{dr}{DR}{Dead Reckoning} 
\newacronym{ins}{INS}{Inertial Navigation Systems}
\newacronym{gps}{GPS}{Global Positioning System} 
\newacronym{lbl}{LBL}{Long Baseline} 
\newacronym{utm}{UTM}{Universal Transverse Mercator}
\newacronym{ned}{NED}{North-East-Down}
\newacronym{enu}{ENU}{East-North-Up}
\newacronym{tf2}{tf2}{The second generation of the Transform Library}
\newacronym{ahrs}{AHRS}{Attitude and Heading Reference System}
\newacronym{ugps}{UGPS}{Underwater \acrshort{gps}}
\newacronym{rmse}{RMSE}{Root Mean Square Error}
\newacronym{uav}{UAV}{Unmanned Aerial Vehicle}
\newacronym{6dof}{6DoF}{Six Degrees of Freedom}
\newacronym{yolo}{YOLO}{You Only Look Once}
\newacronym{DeepSort}{DeepSort}{ Simple Online and Realtime Tracking with a Deep Association Metric}
\newacronym{SORT}{SORT}{ Simple Online and Realtime Tracking}
\newacronym{coco}{COCO}{Common Objects in Context}
\newacronym{taco}{TACO}{Trash Annotations in Context}
\newacronym{kf}{KF}{Kalman Filter}

\begin{document}

\begin{frontmatter}

\title{A BlueROV2-based platform for underwater mapping experiments\thanksref{footnoteinfo}}

\thanks[footnoteinfo]{This work was been financially supported by SeaClear2.0, a project co-funded from the European Union’s Horizon Europe innovation programme under grant agreement No 101093822; and by a grant of the Ministry of Research, Innovation and Digitization, CNCS/CCCDI-UEFISCDI, project number PN-IV-P8-8.1-PRE-HE-ORG-2023-0012, within PNCDI IV.
}

\author[First]{Tudor Alinei-Poiană} 
\author[First]{David Rețe} 
\author[Second]{Davian Martinovici}
\author[Second]{Vicu-Mihalis Maer} 
\author[Second]{Lucian Bu\c{s}oniu}

\address[First]{Technical University of Cluj-Napoca, Romania (e-mail: \{alinei.io.tudor,rete.vi.david\}@student.utcluj.ro)} 
\address[Second]{Technical University of Cluj-Napoca, Romania (e-mail: martinovici.da.davia@student.utcluj.ro, \{vicu.maer,lucian.busoniu\}@aut.utcluj,ro)} %TODO maybe institutional(teams) e-mails for students

\begin{abstract}                % Abstract of not more than 250 words.
We propose a low-cost laboratory platform for development and validation of underwater mapping techniques, using the BlueROV2 \gls{rov}. Both the \gls{rov} and the objects to be mapped are placed in a pool that is imaged via an overhead camera. In our prototype mapping application, the \gls{rov}'s pose is found using extended Kalman filtering on measurements from the overhead camera, inertial, and pressure sensors; while objects are detected with a deep neural network in the \gls{rov} camera stream. Validation experiments are performed for pose estimation, detection, and mapping. The litter detection dataset and code are made publicly available.
\end{abstract}


\end{frontmatter}
%===============================================================================
%===============================================================================
%===============================================================================
%===============================================================================
\section{Introduction}

While autonomous aerial robotics is now commonplace, autonomous underwater robotics is much less so, due to significant extra challenges \citep{towardsAutonomy,Zereik2018}. For example, absolute positioning is much more difficult to obtain, mainly due to the unavailability of global navigation satellite systems; visibility is poor and sonar-based perception not as developed as camera-based; disturbances are stronger, etc. On the other hand, autonomous robotics is sorely needed in underwater applications such as infrastructure inspection or the detection of underwater explosives, so significant efforts are undergoing \citep{Sahoo2019, Zereik2018}. In the SeaClear2.0 project (\url{https://seaclear2.eu}), we are using an autonomous \gls{rov} to detect and map underwater litter, and a second \gls{rov} to collect it.

Field experiments in underwater robotics are logistically, economically, and operationally difficult. Moreover, placing foreign objects in natural underwater environments creates ethical concerns related to the risks of losing them. Thus, early development and testing -- prior and complementary to experiments in real-world scenarios -- should be done in the lab.
%as much work as possible should be done in the lab.

Motivated by all this, we propose here a low-cost laboratory platform for easy development and validation of underwater mapping techniques using an affordable \gls{rov}, the BlueROV2; and a prototype mapping application that uses this platform. The working environment is a pool in which items are placed to be mapped. The pose of the \gls{rov} in the pool is estimated using an \gls{ekf} \citep{Bey_Kalman_filter}, using measurements from an overhead camera (via segmentation of the robot in the image) and from sensors mounted on the \gls{rov}. The \gls{rov} is equipped with an RGB camera, and litter is detected online in the image of this camera using \gls{yolo} \citep{yolov5_performance}, a widely-used deep-learning method for object detection. Mapping is performed with coordinate transforms that rely on the \gls{rov}'s pose. Many of these elements are also being used in SeaClear2.0, so the platform also serves as a testbed for our approach in those field experiments. We provide validation results on detection, pose estimation, and mapping. The dataset to train the network for underwater litter detection, together with our code, are made public at  \url{https://www.kaggle.com/datasets/davianmartinovci/litter-detection} and  respectively \url{https://github.com/ReteDavid/IFAC-2024}.

% This boosts the detection accuracy in comparison to markers [2] and is better suited for real-life experiments where we cannot add a marker to each object. 

\textit{Related work:} \citet{Zereik2018} state that two of the important milestones in this field are localization and mapping in natural, un-prepared environments. \citet{related_3} implement a navigation platform for autonomous underwater vehicles using a trimaran. In the same spirit, we assemble a platform or the development and testing of underwater mapping techniques.

\citet{related_2} present an experimental platform for testing positioning algorithms for unmanned surface vehicles. A marker placed on the dock is detected by a camera, and its relative pose is estimated using the Perspective-n-Point method, leading then to the vehicle pose. While \citet{related_2} use a constant-velocity model coupled with a linear Kalman filter, we use an \gls{ekf} to deal with the nonlinear dynamics of the \gls{rov}. 

\citet{related_5} describe a vision-based underwater platform for position estimation, using again marker detection from camera. The experiments are conducted inside a water tank where the camera floats at one end pointing underwater and the marker is submerged at the other end. Our setup is different in that our positioning camera is above the pool.

\citet{unidu_paper} present a method that is perhaps the closest to ours. They localize the \gls{rov} with respect to a camera mounted on a drone. The GPS coordinates of the drone and the depth given by the \gls{rov}’s pressure sensor are used to estimate its position using the inverse perspective projection method. Here, we additionally estimate the full pose of the \gls{rov} using an \gls{ekf} and then use this estimate for mapping.
% While we use a similar method to estimate the \gls{uuv}’s position, we also apply an \gls{ekf} to further improve the accuracy of the 3D location. 

% As \citet{related_4} state, the \gls{ekf} still remains the simplest, yet decent when it comes to accurate estimation and computationally inexpensive filter when compared to others.  % doesnt' say much so I removed it
\citet{Fallon2013} use a forward-looking sonar to localize objects of interest which are identified by a human operator on a previously obtained acoustic map of the area.
\cite{Massot-Campos} compare methods for underwater 3D reconstruction. According to them, the easiest to assemble and most affordable options for mapping are optical cameras, their main limitations being reduced range and sensitivity to environmental conditions (illumination, turbidity etc.). On the other hand, optical cameras have the advantage of measuring the scene in increased detail and color, which is instrumental for our application - detecting litter objects.

% \vspace{5pt}
The discussion above only touches on a few representative works among the many that perform underwater pose estimation and mapping, and an exhaustive review of the field is beyond our scope. The goal here is not to propose novel algorithms for underwater robotics, but instead to showcase an affordable, easy to use and build laboratory platform for research and development.

\textit{Paper structure:} Platform hardware is presented in Section \ref{sec:platf}, followed by the algorithms used in Section \ref{sec:algo}. We then show validation results in Section \ref{sec:res}, and end with the conclusion and future plans in Section \ref{sec:conc}. 
% -------------------------------------------------------------------------
% ----- SECTION BREAK -------------------------------
% ------------------------------
\section{Hardware components}\label{sec:platf}
%Hardware description
The hardware components constituting the experimental platform include an overground pool measuring \qtyproduct{2.7 x 5.5 x 1.2}{\metre} with a water depth of $1$\,m, an overhead camera, an underwater \gls{rov}, and a topside computer running the \gls{ros}. % More details will be given next.

\begin{figure}[!tbh]
    \centering
    \includegraphics[width=0.8\linewidth]{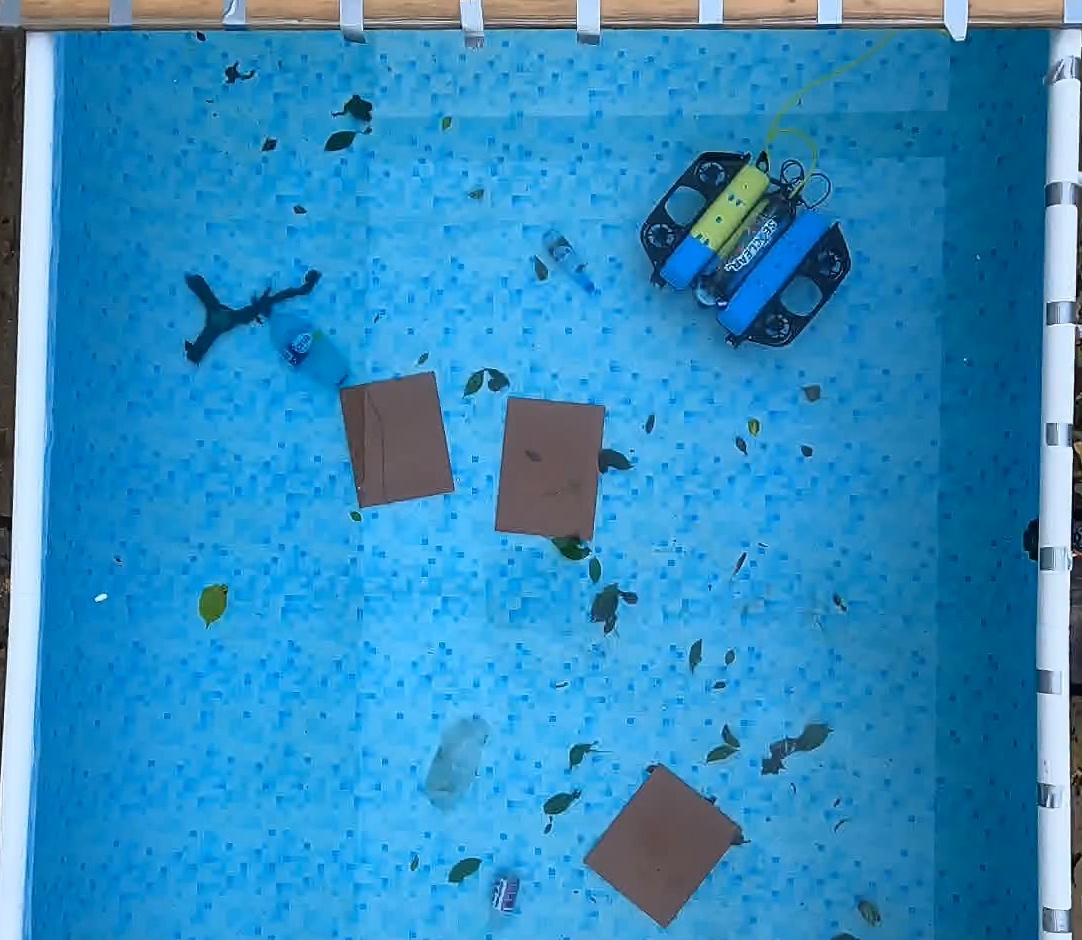}
    \caption{Capture of working area from the overhead camera}
    \label{fig:overheadView} 
\end{figure}

\textit{\gls{rov}}:
A BlueROV2 is used, manufactured by the BlueRobotics Inc.\ company.
It is an inspection class vehicle capable of diving up to \qty{100}{\metre}, powered by a Li-ion battery, and communicating to the surface through a tether connected to a proprietary networking interface. The application-relevant sensors on the BlueROV2 are an \gls{imu} which alongside a magnetometer provide attitude measurements, a pressure sensor used for depth/altitude measurement and a camera allowing a \ang{90} range of tilt. For operation in lower lighting conditions, lights are also fitted to the front of the vehicle. The \gls{rov} can be seen in Figure \ref{fig:overheadView}, top-right.
% Figure \ref{fig:bluerov2} shows the BlueROV2.
% \begin{figure}[!tbh]
%     \centering
%     \includegraphics[width=\linewidth]{figures/bluerov2.png}
%     \caption{The BlueROV2}
%     \label{fig:bluerov2} 
% \end{figure}

\textit{Overhead camera}:
A GoPro HERO 9 camera is used for the overhead view. It is mounted above the pool such that field of view includes the working area; through a cable connection, it streams 1080p/30fps video to the topside computer. The intrinsic camera parameters where obtained through a checkerboard calibration. Figure \ref{fig:overheadView} shows a capture from one of the experiments produced by this camera.

\textit{Topside computer}:
A consumer-grade notebook PC is connected to the BlueROV2's networking interface, enabling control of the \gls{rov}, and to the overhead camera. Data from the \gls{rov} and the overhead camera are fed into the \gls{ros} network. After this, the data are consumed by the detection, mapping and positioning algorithms running on the same computer.
%Diagram

% -------------------------------------------------------------------------
% ----- SECTION BREAK -------------------------------
% ------------------------------
\section{Algorithmic components}\label{sec:algo}

\begin{figure}[!tbh]
    \centering
    \includegraphics[scale=0.4]{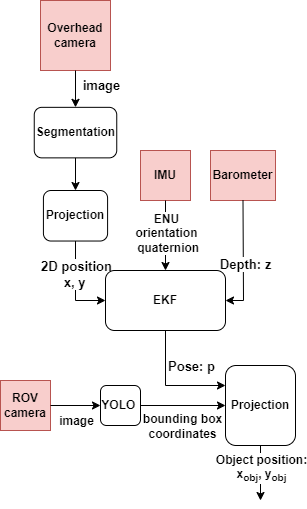}
    \caption{Functional diagram}
    \label{fig:functional_diagram} 
\end{figure}

The algorithmic workflow is presented in Figure \ref{fig:functional_diagram}. Segmentation in the overhead image is used to find a yellow patch on the \gls{rov}, whose coordinates are  then used to determine the planar coordinates of the the \gls{rov}. The \gls{imu} and magnetometer are fused using an internal filter in the ArduSub firmware in order to provide both orientation and angular velocity, while depth is measured using the barometer. These data are then fused using an (external) \gls{ekf}, resulting in a complete pose for the robot. Simultaneously, the \gls{rov}'s internal camera captures underwater images which are processed using the \gls{yolo} detection algorithm. The resulting bounding box coordinates are transmitted to the object mapping algorithm, which, using the \gls{ekf} pose estimate, maps the detected litter items.

% -------------------------------------------------------------------------
% ----- SUBSECTION BREAK -------------------------------
% ------------------------------
\subsection{Detection}\label{yolo}
The detection algorithm is used to process the visual feed coming from the BlueROV2 camera in order to recognize litter in the water. It then highlights each detected object through a bounding box. We use the fifth version of the "You Only Look Once" deep-learning object detection model, YOLOv5, which is popular in computer vision tasks due to its fast and light design \citep{yolov5_performance}. The algorithm undergoes training on one dataset, followed by testing on a separate validation dataset. To increase its accuracy, we train the model multiple times, each time starting from a previously finished training checkpoint in order. 

\begin{figure}[!tbh]
	\centering
	\includegraphics[scale=0.25]{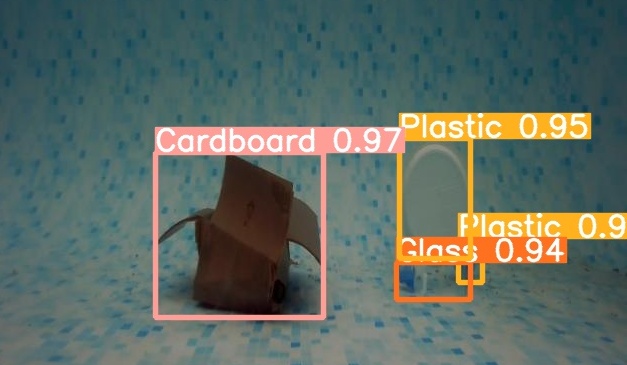}
	\caption{Example of YOLOv5 detections in an \gls{rov} camera frame}
	\label{fig:yolo_detection}
\end{figure}

Our datasets include images depicting litter on land and underwater from three different sources: underwater photos captured using the BlueROV2 camera, on-land photos captured by phone, and a publicly available dataset called ``trash annotations in context'' \citep{taco}. The training dataset has 3194 images, the validation dataset has 1079 images, and the testing dataset has 1000 images. Each image can contain one or more objects of the required classes: plastic, metal, cardboard and glass, as illustrated in Figure \ref{fig:yolo_detection}. In the training dataset, each class except plastic has approximately 1000 labeled objects, while the plastic class has approximately 2500. Prior to using an image, it is resized to a $640\times640$ resolution. The images and the annotations used for training, validation, and testing are publicly available at \url{https://www.kaggle.com/datasets/davianmartinovci/litter-detection}.

To simulate different environments with varying light conditions and the possibility of murky water, augmentations provided by the Roboflow framework \citep{roboflow} were used, such as random brightness adjustments, blur, \ang{90} rotations and random pixel noise up to 3\%. The model underwent training for 300 epochs in 5 iterations.

To minimize identity switches in frames, the DeepSort \citep{deepsort} algorithm is used to track multiple objects in subsequent frames, especially during periods of occlusion. % DeepSort utilizes a Kalman filter for frame-by-frame data association. 

\subsection{Segmentation}\label{segmentation}

To determine the position of the BlueROV2 in the overhead camera image, we segment a yellow patch attached to its top. A thresholding procedure is applied to identify the yellow regions in the image, in the HSV (hue, saturation, value) color space. %In this space the image has 3 characteristics: hue, representing the color type; saturation, the amount of white present in color; value, the brightness of the color. 
First, the intervals were selected so the yellow patch would be detected in optimal light conditions. Subsequently, through trial and error experiments involving changing light conditions, the intervals were adjusted. The resulting intervals are $H\in[20, 50]$, $S\in[98, 255]$ and $V\in[115, 255]$. To filter out smaller accidental detections of e.g.\ water reflections, we select the largest region found. This region's centroid, which is the average of all the pixel coordinates in the region, will indicate the BlueROV2 position within the camera frame.

\subsection{Coordinate transform}\label{coord_transform}

The next component determines both the real-world position of the BlueROV2 from the segmented region's centroid; as well as the real-world positions of litter objects from the centers of the bounding boxes around YOLOv5 detections. 

We use the pinhole camera model, which describes the link between a point $\mathtt{P_w}$ in 3D and its projection $\mathtt{x, y}$, represented in homogeneous coordinates in the image plane:\footnote{Notations in this subsection use a different font to distinguish them from pose estimation notations, while still keeping standard symbols in both settings.} 
\begin{gather}\label{eq4}
 \mathtt{[x, y, 1]^\top
 = K\cdot T \cdot P_w}
\end{gather}
where $\mathtt{K}$ is the camera intrinsic matrix (obtained via calibration using checkerboards); and $\mathtt{T}$ is the transformation matrix (rotation, translation) between the world and camera frames.
\begin{figure}[h]
    \centering
    \includegraphics[width=0.9\linewidth]{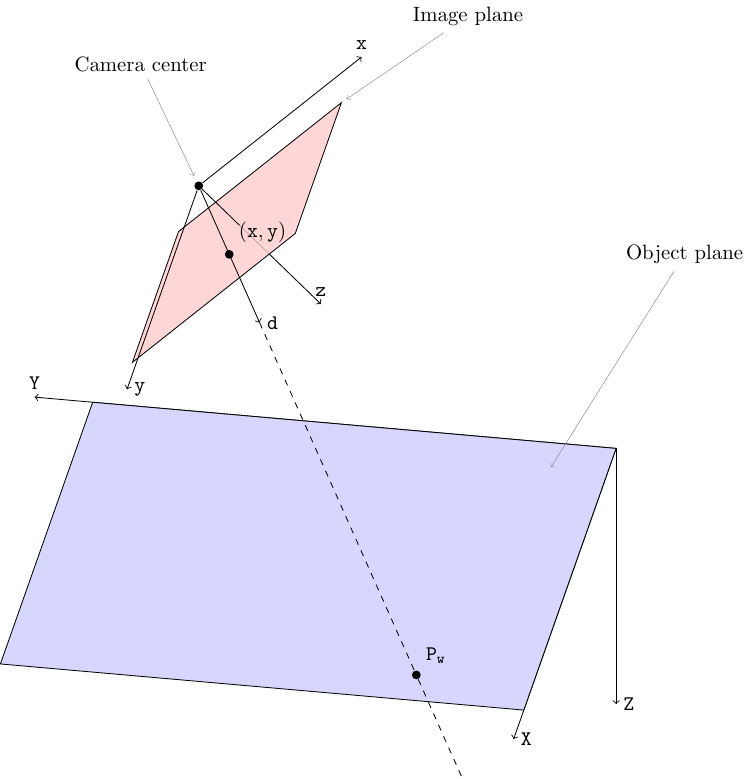}
    \caption{Procedure to find an object's world-frame coordinates}
    \label{fig:plane_pose} 
\end{figure}

The procedure relies on inverse projection as described by \citet{maer}, and is illustrated in the Figure \ref{fig:plane_pose}. We begin with the pixel coordinates $\mathtt{x,y}$ of the region centroid or the center of the detection bounding box, and multiply these coordinates with the inverse of the camera intrinsic matrix $\mathtt{K}$ to obtain a new set of coordinates that provide the direction $\mathtt{d}$ towards the object, in the camera reference frame. This line goes through the center point of camera $\mathtt{C=[0,0,0]^T}$ and intersects both the image of the detected object as well as the object itself. Next, we transform the camera-frame direction $\mathtt{d}$ to a world-frame direction $\mathtt{d_w}$ by multiplying with the inverse of the rotation matrix between the camera and world frames $\mathtt{R}$:
 \begin{equation}
    \mathtt{d_w = R^{-1} \cdot d}
\end{equation}
% Continuing, we will keep the notations for the coordinates even though homogeneous coordinates are only used for the initial step using the inverse intrinsic matrix.

The world-frame position of the object is given by the intersection of $\mathtt{d_w}$ and the horizontal plane in which the object lies. The world coordinate $\mathtt{Z_{w,obj}}$ of this plane is known for both the BlueROV2 (the depth given by the pressure sensor), and for the litter objects ($\mathtt{Z_{w,obj}=-1}$, at the bottom of the pool, which is 1\,m deep). Note that $\mathtt{Z_w = 0}$ at the water surface. The object's depth gives us the additional necessary information required by the camera perspective projection to compute the 3D position of the object from the 2D position in the frame. Further, we require the camera poses: the fixed overhead camera is situated at $\mathtt{C_w = [0,0,1.2]^T}$ oriented downwards, and the camera mounted on the BlueROV2 has a variable pose computed using the \gls{ekf} pose estimate and a static transform that takes into account the camera's position relative to the \gls{rov} center, and the camera's tilt. 

Finally, the coordinates $X, Y$ of the object in real world (belonging to $\mathtt{P_w}$), as well as $\mathtt{s}$, the distance along direction $\mathtt{d_w}$ until the object is reached, are determined by solving the system of equations:
\begin{equation}
    \begin{cases}
        \mathtt{P_w = C_w + d_w \cdot s}\\
        \mathtt{\begin{bmatrix}
            0&0&1
        \end{bmatrix} \cdot P_w = Z_{obj}}
    \end{cases}
\end{equation} 
% \textcolor{red}{, in this application, East-North-Up frame was preferred in order to comply with the ROS standard.}
The coordinates obtained are transformed into the East-North-Up frame, a local Cartesian coordinate system with its axes pointing eastward along the parallel of latitude, northward along the meridian, and upward. This frame was chosen since it is standard in \gls{ros}, although in marine engineering the typical coordinates frame is North-East-Down.

\subsection{Pose estimation}

In this subsection we introduce the algorithm used for the \gls{rov}'s pose estimation, the extended Kalman filter (\gls{ekf}) \citep{Bey_Kalman_filter}. The \gls{ekf} performs nonlinear state estimation by linearizing at each step the following non-linear model that describes the movement of a robot:
\begin{equation}
    \begin{cases}
        x_k=f(x_{k-1})+w_{k-1}\\
        y_k=h(x_{k})+v_{k}\\
    \end{cases}
    \label{eqn:rigid_body_model_eq}
\end{equation}
where $k$ is the discrete time step, $f(x_k)$, called transition function, describes the motion equations of the robot, $x_k$ is the robot's state and $v_k$ is the process noise. In the second equation $y_k$ is the measurement vector, the sum between the measurement equation $h(x_k,u_k)$ and measurement noise $w_k$. The discrete time step is denoted by the $k$ subscript. Both $v_k$ and $w_k$ are assumed to be zero-mean Gaussian white noise. 

A rigid body motion model explained in %\url{https://control.asu.edu/Classes/MMAE441/Aircraft/441Lecture9.pdf}
%\citep{rigid_body2} 
\citep{rigid_body_kin}
is used to describe the \gls{rov} in \eqref{eqn:rigid_body_model_eq}. This model does not include any hydrodynamic effects or thruster input, relying instead solely on a double integrator with constant acceleration, for which the observability condition is fulfilled whenthe full pose is measured. The state $x$ consists of: 
\[
(X, Y, Z, \phi, \theta, \psi, \dot{X}, \dot{Y}, \dot{Z}, \dot{\phi}, \dot{\theta}, \dot{\psi}, \ddot{X}, \ddot{Y}, \ddot{Z})
\]
where $(X,Y,Z)$ is the \gls{rov} position, $(\phi, \theta, \psi)$ is the orientation described using Euler angles (roll, pitch, and yaw, respectively), $(\dot{X}, \dot{Y}, \dot{Z})$ are linear velocities, $(\dot{\phi}, \dot{\theta}, \dot{\psi})$ the angular velocities and $(\ddot{X}, \ddot{Y}, \ddot{Z})$ the linear accelerations. As mentioned, in our case, the system does not have an input $u$. Note that the accelerations do not integrate jerk signals, instead they just integrate noise.

The \gls{ekf} involves two main steps: prediction and update. In the prediction step:
\begin{equation}
    \begin{cases}
        A_k=\left. \frac{\partial f}{\partial x}\right|_{(\hat{x}_{k-1})}\\
        P_{k|k-1}=A_kPA_k^\top+Q\\
        x_{k|k-1}=f(\hat{x}_{k-1})\\
    \end{cases}  
    \label{prediction_ekf}
\end{equation}
the system is linearized by computing the Jacobian of $f$ with respect to $x$ and $u$ and evaluating it using the previous state estimate $\hat{x}_{k-1}$ and input $u_{k-1}$, which leads to the state matrix $A_k$ at step $k$. Then, the estimator predicts the error covariance matrix $P_{k|k-1}$ and the new system state $x_{k|k-1}$. The process noise covariance matrix $Q$ can be seen as a way of declaring confidence in the accuracy of the model. Usually, $Q$ is diagonal, with larger values on the diagonal indicating smaller confidence in the model for that state. % Larger variances suggest that the model might not accurately reflect the dynamics of the actual system.
The experimentally tuned variances in $Q$ for our rigid-body model are (see above for the order of the states): 0.1, 0.1, 0.06, 0.03, 0.03, 0.06, 0.025, 0.025, 0.04, 0.01, 0.01, 0.02, 0.01, 0.01 and 0.015.

The update step of \gls{ekf} is:
\begin{equation}
        \begin{cases}
            C_k=\left. \frac{\partial h}{\partial x}\right|_{(x_{k|k-1})}\\
            K_k=P_{k|k-1}{C_k}^\top(C_kP_{k|k-1}C_k^\top+R)^{-1}\\
            \hat{x}_k=x_{k|k-1}+K_k(y_k-C_kx_{k|k-1})\\
            P=(I-K_kC_k)P_{k|k-1}(I-K_kC_k)^\top+K_kRK_k^\top
        \end{cases}
        \label{eqn:update_ekf}
\end{equation}
where the measurement matrix $C_k$ at $k$ is found by linearizing and evaluating the measurement transition function $h$ at $x_{k|k-1}$, $K_k$ is the Kalman gain at the current step, and $R$ is the measurement covariance matrix. Assuming that the measurements are not correlated, $R$ is a diagonal matrix with the variances of each measurement on the diagonal. The larger the variance, the less confidence the filter will have in that measurement, resulting in a smaller update gain in $K_k$. The actual state update (correction)  is done in the third equation, leading to $\hat{x}_k$. In the final step, the error covariance matrix $P$ is updated.

When applied to the BlueROV2, the filter measurements include orientation and angular velocities determined by the \gls{imu} and magnetometer, the planar coordinates measured via the overhead camera, and the depth provided by the internal barometer. The standard deviations for the linear accelerations and angular velocities were taken from the technical sheet of the \gls{imu}, and are respectively 0.01 $\si{\metre}/\si{\sec}^2$ and 0.04 $\si{\deg}/\si{\sec}$. 
%Since the orientation with respect to the global reference system involves nonlinear combinations of measurements from the magnetometer and linear accelerations, a
An experimental procedure was employed to determine the standard deviation of the orientation. The orientation signal (Euler angles) was filtered, and the difference between the original signal and the filtered version provided an approximation of the noise. The root mean square of this approximation is the standard deviation used in the \gls{ekf}. The same procedure was performed for the camera's $XY$ measurements and the barometer's depth $Z$, leading to:
\begin{align*}
\sigma_{\phi} &= 0.0018\,\si{\radian} & \sigma_{\psi} &= 0.0003\,\si{\radian} & \sigma_{\gamma} &= 0.0054\,\si{\radian} \\
\sigma_{X} &= 0.012\,\si{\metre} & \sigma_{Y} &= 0.08\,\si{\metre} & \sigma_{Z} &= 0.0024\,\si{\metre}
\end{align*}
where $\sigma_q$ denotes the standard deviation of quantity $q$.

\subsection{Mapping}
\label{mapping_subsection}
We explain now how we integrate \gls{yolo} from Section \ref{yolo}, segmentation from Section \ref{segmentation}, and the inverse projection method from Section \ref{coord_transform}, in order to perform the mapping of the litter objects; see again Figure \ref{fig:functional_diagram}. Segmentation is used to detect the BlueROV2, whose position is then computed using the inverse projection method.
% The litter cannot be detected simply based on color, therefore a method that detects the special features of an object in the image is proposed. 
Using the \gls{yolo} method while navigating, the litter is detected by the camera mounted on the BlueROV2, which gives the coordinates of the bounding box enclosing the litter object. We take into account the variable pose of the \gls{rov} camera in the real world, which is inferred from the pose of the vehicle as found by the \gls{ekf} and the camera's static transform. The real-world coordinates of the litter are then determined by applying the inverse projection method on the bounding box center. Each litter item found in this way is marked on a 2D map, stored as an array of litter coordinates. 
To avoid marking duplicate litter objects for detections in each frame, if the distance between a new detection and the closest existing object is below a certain threshold (chosen here as 30 \SI{}{\centi\metre}), a new object is not added.

% To validate these coordinates, we perform a manual detection (click on the litter's center) from the overhead camera. 

The repository containing the code used for the experiments can be found at \url{https://github.com/ReteDavid/IFAC-2024}.

% -------------------------------------------------------------------------
% ----- SUBSECTION BREAK -------------------------------
% ------------------------------

% -------------------------------------------------------------------------
% ----- SECTION BREAK -------------------------------
% ------------------------------
\section{Experimental results}\label{sec:res}
Next, the validation procedures and results for localization, detection and litter mapping will be presented. First, we independently validated the robot's pose estimate in a smaller indoor tank, and the object detection algorithm on the validation dataset. After that, we performed the validation of litter mapping in the larger pool, in which we placed litter objects of interest (plastic bottles, cartons etc.). The \gls{rov} is manually driven for these experiments.

\begin{figure}[!tbh]
    \centering
    \includegraphics[scale=1]{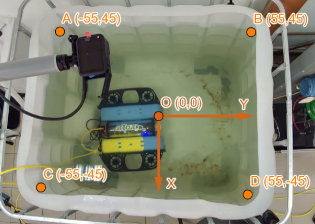}
    \caption{Experimental set-up for pose validation}
    \label{fig:setup_bazin_mic} 
\end{figure}
\begin{figure}[!tbh]
    \centering
    \includegraphics[scale=0.35]{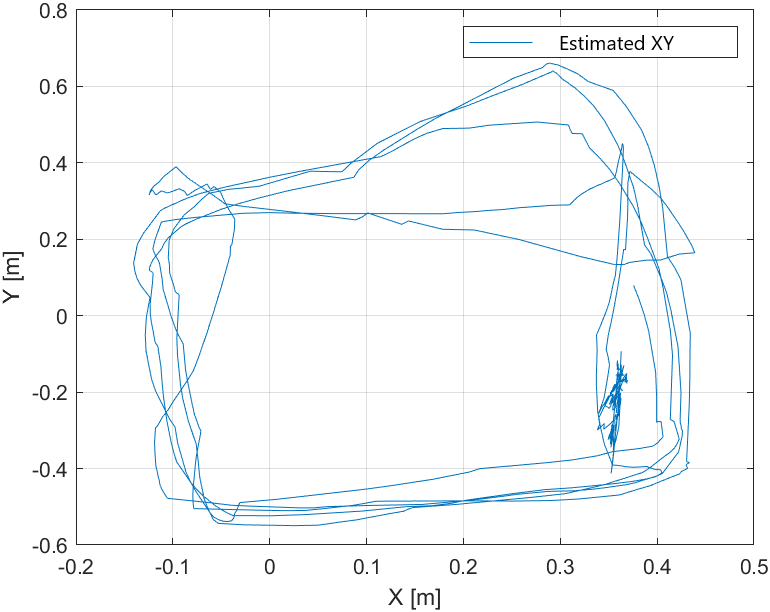}
    \caption{Planar position estimate in the tank}
    \label{fig:rezultate_est_camera_4ture} 
\end{figure}

The indoor validation for the \gls{rov}'s pose was performed in a small-sized water tank of \qtyproduct{1 x 1 x 1}{\metre}. The origin is taken at the center of the tank, on the water surface. The corners of the basin are located at (55,45), (-55,45), (55,-45), and (-55,-45) \SI{}{\centi\metre}, as shown in Figure \ref{fig:setup_bazin_mic}. Pose estimation was performed with the \gls{ekf} while the \gls{rov} was driven approximately around the edges of the tank, and the results are shown in Figure \ref{fig:rezultate_est_camera_4ture}. The real coordinates of the \gls{rov} were measured with a measuring tape in the corners of the tank, taking into account the length and width of the \gls{rov}. Then, the root mean squared error (RMSE) between these coordinates and those provided by the pose estimate in the tank corners was computed. The resulting errors are $\mathrm{RMSE}_X = \SI{5.5}{\centi\meter}$ and $\mathrm{RMSE}_Y = \SI{8.7}{\centi\meter}$ for the two coordinates. By using the barometer we were able to  determine the robot's depth with good accuracy (significantly better than for the planar position, on the order of 1cm; results not shown here). Overall, results are satisfactory.
% \begin{figure}[!tbh]
%     \centering
%     \includegraphics[scale=0.35]{figures/adancime_mas_vs_estimata2.png}
%     \caption{Depth estimation}
%     \label{fig:depth_estimation_comparison} 
% \end{figure}\\
% The position of the ROV may exhibit jumps due to the loss of color detection for the yellow contour placed on the ROV's surface.
% In Figure \ref{fig:rezultate_est_camera_4ture} we can see a comparison between the measured and estimated $Z$ coordinate.

\begin{figure}[!tbh]
	\centering
	\includegraphics[width=\columnwidth]{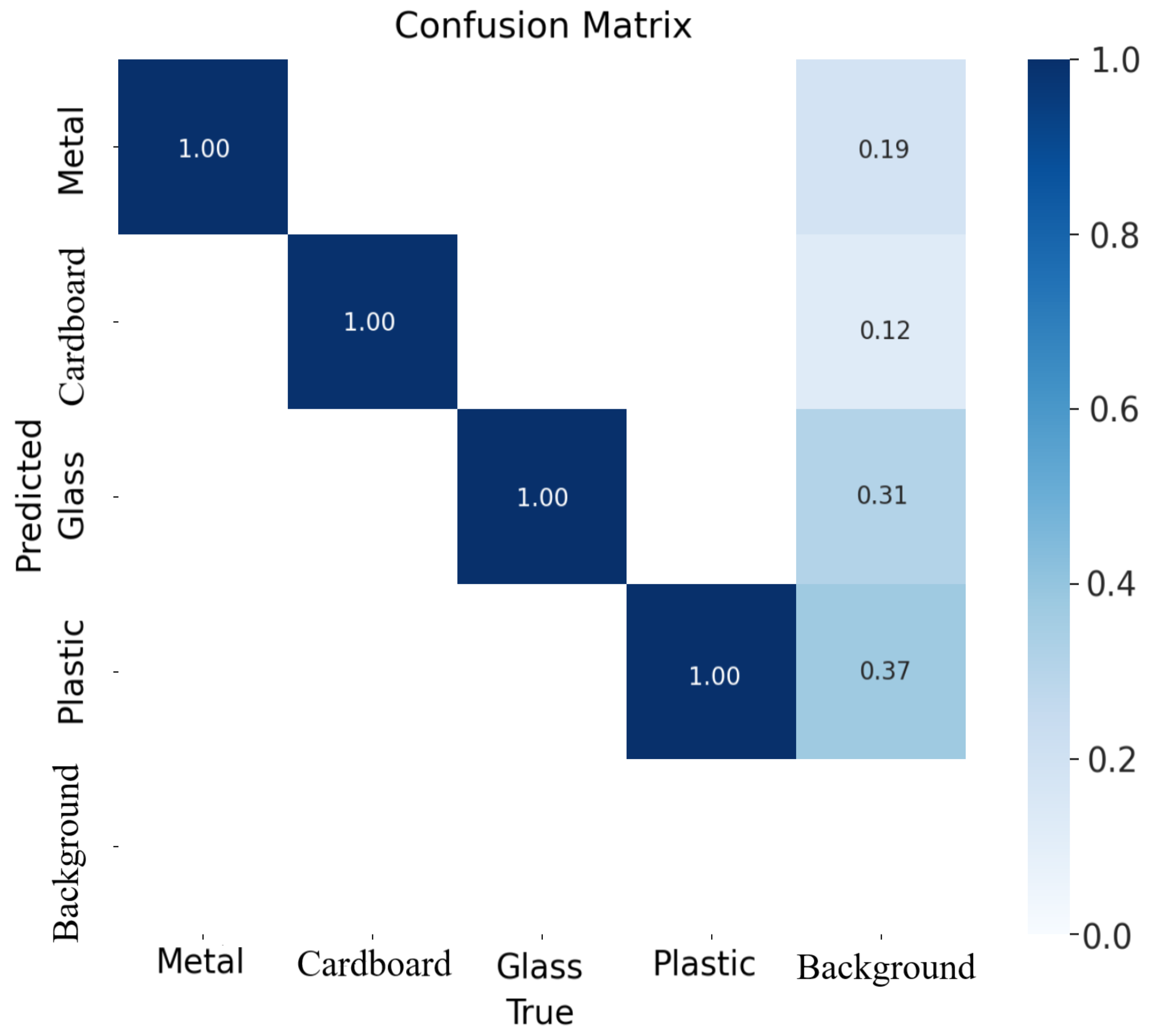}
	\caption{Confusion matrix for the trained YOLOv5 model}
	\label{fig:confusion_matrix}
\end{figure}

Moving on to the validation of the YOLOv5 detector, Figure \ref{fig:confusion_matrix} displays the confusion matrix for the final trained model on the validation data, for the $N=4$ classes of interest: plastic, glass, carton, and metal. In this $N+1\times N+1$ matrix, the element at position $i, j$ represents the ratio of predicted class instances in row $i$ to the actual instances in column $j$ (with an extra row and column for the background). The main diagonal represents correctly classified instances, while other values indicate instances where the model misclassified objects. In our case, the four targeted classes are identified very well, but the model sometimes has false classifications of background as litter objects, probably because of similarities between said objects and background.

Having validated our estimator and detector, we now bring together the complete litter mapping framework from Figure \ref{fig:overheadView}, and test it in the larger pool. Ground-truth positions of the litter objects were manually found with the help of two grids placed on the sides of the pool, and the RMSE was computed between these positions and the locations mapped using the procedure of Section \ref{mapping_subsection}. The resulting errors for both $X$ and $Y$ coordinates are 12 \SI{}{\centi\meter}, and their sources are (i) the pose estimation error, and (ii) the fact that the ground-truth position is in the real center of the object, while the bounding box predicted using \gls{yolo} is sometimes off-center.

\begin{figure}[!tbh]
    \centering
    \includegraphics[scale=0.2]{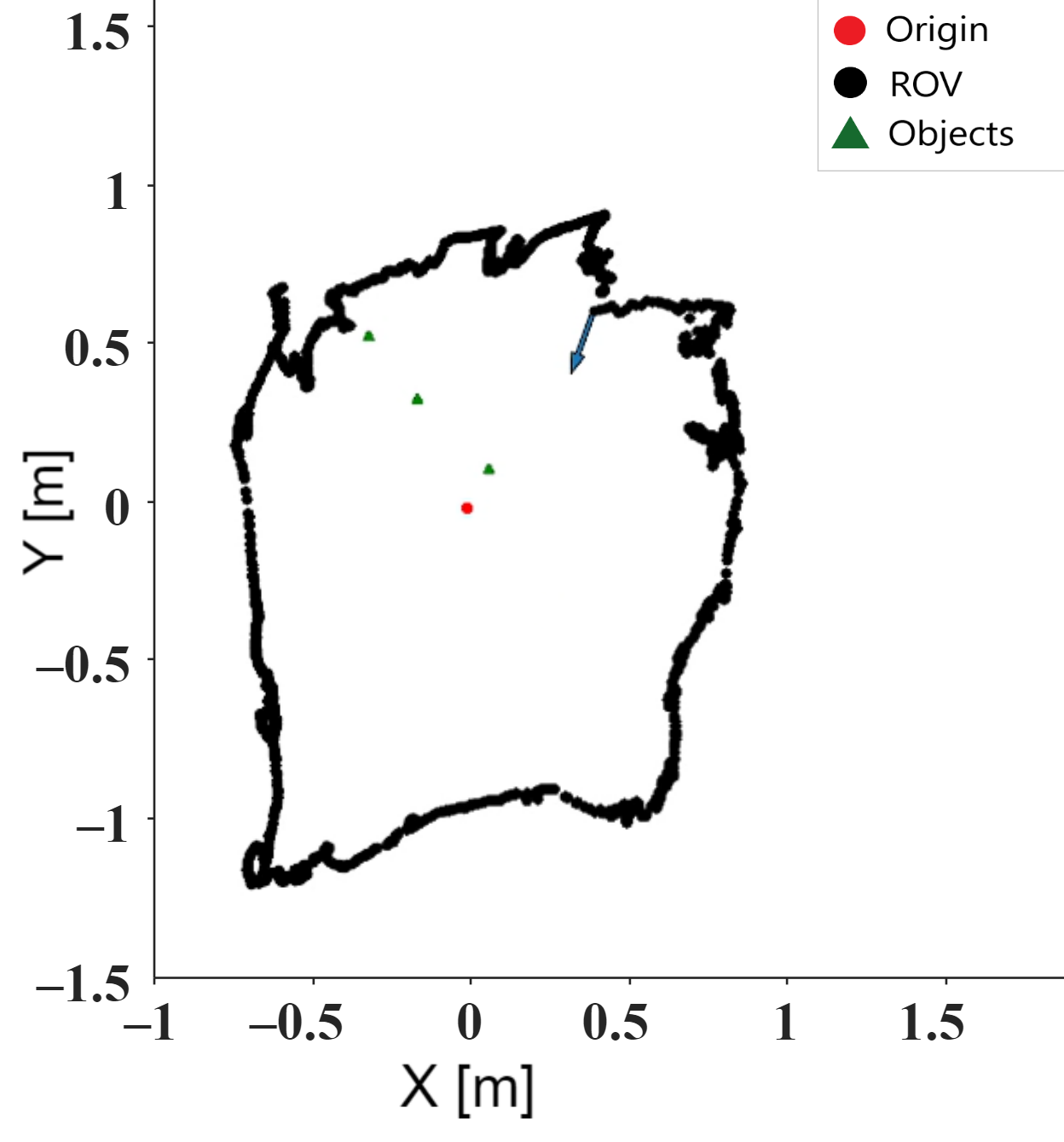}
    \caption{Pose estimation and object mapping results in the final experiment}
    \label{fig:pose_estim_obj_mapping} 
\end{figure}

Figure \ref{fig:pose_estim_obj_mapping} illustrates object mapping alongside \gls{ekf} estimation, where the black dotted line represents the history of the \gls{rov}'s position, the red dot represents the center of the pool, which is the origin of our coordinate frame, the blue arrow indicates the robot's current yaw, and the green triangles depict the position of the mapped litter items.\footnote{We removed some unreliable detections made with very large yaw errors due to magnetometer issues.} 
%\begin{figure}[!tbh]
%    \centering
%    \includegraphics[scale=0.23]{figures/big_basin_coords.png}
%   \caption{Outside pose estimation results}
%    \label{fig:outside_pose_est} 
%\end{figure}

% \begin{comment}
   % One observation about the object mapping is that, though the objects are mapped correctly, multiple mapped objects can be displayed for only one real item. This happens because each object receives a set of multiple measurements and bounding box coordinates and if the bounding box coordinates change only for a bit, a new mapped object will appear on the map. This could be solved heuristically by computing the distance between the detected objects and the newly received item and removing the mapped object if the distance is less than a threshold. This method works well on the assumption that all the objects have at least that threshold distance between them.\\ 
% \end{comment}
% \\
For a visual presentation of the entire pipeline and the final experiment discussed above, see the video at \url{https://www.youtube.com/watch?v=DHD8hmEt5_U}.

% -------------------------------------------------------------------------
% ----- SECTION BREAK -------------------------------
% ------------------------------
\section{Conclusions}\label{sec:conc}
In this paper, we have described a platform for the development and experimental  validation of underwater mapping algorithms. Future improvements in pose estimation include the integration of a true dynamical model in the \gls{ekf}, incorporating e.g.~hydrodynamics and sensor bias; as well as finding yaw angles in the overhead camera image. Options for future hardware extensions include a sonar, a Doppler velocity logger to improve pose estimation, or a manipulator to pick up litter. Quantitative comparisons between different pose estimators or object detectors will be performed; and self-localization and mapping can be attempted.
%\begin{ack}
%Place acknowledgments here.
%\end{ack}

{\small 
\bibliography{main}            
}% bib file to produce the bibliography
% with bibtex (preferred)

\end{document}